\documentclass[sigconf]{acmart}

\usepackage{graphicx}
\usepackage{multirow}
\usepackage{amsmath}
\usepackage[table]{xcolor}
\usepackage[utf8]{inputenc} 
\usepackage[T1]{fontenc}    
\usepackage{hyperref}       
\usepackage{url}            
\usepackage{booktabs}       
\usepackage{amsfonts}       
\usepackage{nicefrac}       
\usepackage{microtype}      
\usepackage{xcolor}         

\usepackage{wrapfig}  
\AtBeginDocument{%
  }

\begin{document}

\title{QuiZSF: A Retrieval-Augmented Framework for Zero-Shot Time Series Forecasting}


\author{Shichao Ma}
\affiliation{%
  \institution{University of Science and Technology of China}
  \city{Hefei}
  \country{China}
}
\email{mashichao@mail.ustc.edu.cn}

\author{Zhengyang Zhou}
\authornote{Dr. Zhengyang Zhou and Prof. Yang Wang are corresponding authors.}
\affiliation{%
  \institution{University of Science and Technology of China}
  \city{Hefei}
  \country{China}
}
\affiliation{%
  \institution{Suzhou Institute for Advanced Research, USTC}
  \city{Suzhou}
  \country{China}
}
\email{zzy0929@ustc.edu.cn}

\author{Qihe Huang}
\affiliation{%
  \institution{University of Science and Technology of China}
  \city{Hefei}
  \country{China}
}
\email{hqh@mail.ustc.edu.cn}

\author{Binwu Wang}
\affiliation{%
  \institution{University of Science and Technology of China}
  \city{Hefei}
  \country{China}
}
\email{wbw2024@ustc.edu.cn}


\author{Yang Wang}
\authornotemark[1] 
\affiliation{%
  \institution{University of Science and Technology of China}
  \city{Hefei}
  \country{China}
}
\affiliation{%
  \institution{Suzhou Institute for Advanced Research, USTC}
  \city{Suzhou}
  \country{China}
}
\email{angyan@ustc.edu.cn}

\renewcommand{\shortauthors}{Shichao Ma, Zhengyang Zhou, Qihe Huang, Binwu Wang, and Yang Wang}

\begin{abstract}
Accurate forecasting of sequential data streams is a cornerstone of modern Web services, supporting applications such as traffic management, user behavior modeling, and online anomaly prevention. However, in many Web environments, new domains emerge rapidly and labeled history data is scarce, which makes zero-shot forecasting particularly challenging. Existing time-series pre-trained models (TSPMs) show promise but they lack the ability to dynamically incorporate external knowledge, while conventional retrieval-augmented generation (RAG) methods are rarely extended beyond text. In this work, we present \textbf{QuiZSF}, a retrieval-augmented forecasting framework that integrates search and forecasting for time series data. The framework performs search by retrieving structurally similar sequences from a large-scale time-series database, and it performs forecasting by integrating the retrieved knowledge into the target sequence. Specifically, QuiZSF introduces a \textbf{ChronoRAG Base}, a hierarchical tree-structured database that enables scalable and domain-aware retrieval, a \textbf{Multi-grained Series Interaction Learner} that captures fine- and coarse-grained dependencies between target and retrieved sequences, and a \textbf{Model Cooperation Coherer} that adapts retrieved knowledge to TSPMs. This design teaches models to actively perform search, align auxiliary information across modalities, and leverage it for more accurate forecasting. Extensive experiments on five public benchmarks demonstrate that QuiZSF consistently outperforms strong baselines, ranking first in up to \textbf{87.5\%} of zero-shot forecasting settings while maintaining high efficiency. 
\end{abstract}




\maketitle

\section{Introduction}

Modern Web services rely heavily on sequential data streams, ranging from traffic monitoring and user behavior modeling to anomaly prevention in large-scale online platforms. Accurate forecasting of such time series is therefore a cornerstone for building reliable and intelligent Web applications. However, Web environments are highly dynamic and data-scarce: new domains and conditions frequently emerge, while historical labels are limited or unavailable. These challenges make \textit{zero-shot time-series forecasting} (ZSF), which aims to infer future trends of unseen sequences without domain-specific supervision, a crucial yet difficult problem~\cite{wu2022timesnet, gruver2024large, jin2023time, ekambaram2024tiny}.

Recent progress in \textit{time series pre-trained models} (TSPMs) has shown promising results for ZSF by transferring knowledge across datasets, analogous to pre-training in NLP and CV. TSPMs can be categorized into two groups~\cite{ekambaram2024tiny}: (i) numerical Non-LLM based TSPMs such as Moment~\cite{goswami2024moment}, TTM~\cite{ekambaram2024tiny}, TimesFM~\cite{das2024decoder}, and Moirai~\cite{woo2024unified}, and (ii) textual LLM based models that reconstruct language models for time series forecasting, such as LLMTime~\cite{gruver2024large}, Time-LLM~\cite{jin2023time}, and GPT4TS~\cite{zhou2023one}. Although both categories achieve strong forecasting accuracy, they face two critical limitations. First, real-world sequences are continually evolving, yet TSPMs cannot efficiently incorporate new knowledge without costly fine-tuning~\cite{tan2024language}. Second, time series often exhibit structural similarity across domains~\cite{tire2024retrieval}, but TSPMs lack mechanisms to retrieve and reuse such auxiliary patterns to improve generalization.

\begin{figure}[ht]
    \centering
    \includegraphics[width=0.45\textwidth]{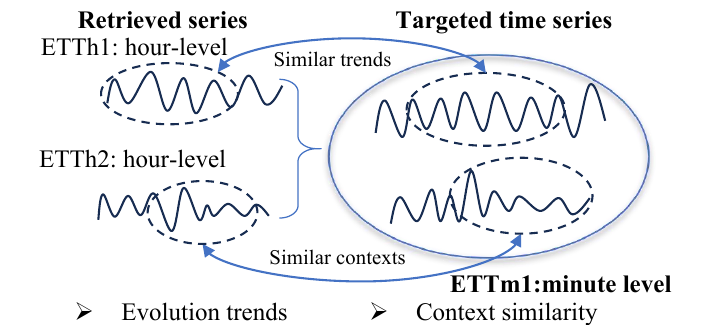}
    \caption{Motivation. Time series across domains often exhibit similar temporal patterns, which can be retrieved and reused as auxiliary knowledge.}
    \label{fig:motivation}
\end{figure}

To address these limitations, we propose integrating TSPMs with retrieval-augmented generation (RAG), a paradigm that has shown great success in natural language processing but remains underexplored for time series forecasting. In this approach, a search module first retrieves relevant sequences from a large-scale temporal database, and a forecasting module then integrates the retrieved knowledge into the target sequence prediction process.
As illustrated in Figure~\ref{fig:motivation}, retrieved auxiliary sequences can provide valuable contextual signals such as periodicity, seasonal shifts, or abrupt transitions, which enhance forecasting quality and reduce hallucinations in zero-shot scenarios.

Nevertheless, combining retrieval with forecasting introduces three core challenges:
(i) \textbf{Efficient and domain-sensitive storage and retrieval}: indexing millions of sequences for fast and relevant cross-domain search remains difficult at Web scale;
(ii) \textbf{Multi-level feature extraction}: retrieved sequences vary in scale, domain, and noise, requiring expressive yet lightweight interaction modeling;
(iii) \textbf{Modality-aligned representation integration}: numerical Non-LLM based TSPMs demand feature-level fusion, while textual LLM-based models require structured prompts that translate retrieved sequences into language-compatible inputs.

To address these challenges, we propose the \textbf{Qui}ck \textbf{Z}ero-shot time-series \textbf{S}earch and \textbf{F}orecasting framework (\textbf{QuiZSF}). QuiZSF is a retrieval-augmented forecasting framework that integrates scalable storage, cross-series interaction learning, and modality-aware adaptation. Our main contributions are summarized as follows:
\begin{itemize}
    \item We build the \textbf{ChronoRAG Base (CRB)}, a hierarchical tree-structured temporal database, together with a \textbf{Hybrid and Hierarchical TS Retrieval (HHTR)} strategy for fast and domain-sensitive search.
    \item We design the \textbf{Multi-grained Series Interaction Learner (MSIL)}, a module that captures fine-grained dependencies and coarse-grained trends from retrieved sequences to improve target sequence understanding.
    \item We develop the \textbf{Model Cooperation Coherer (MCC)}, a dual-branch adapter that integrates retrieved information into both Non-LLM and LLM-based TSPMs.
    \item We conduct extensive experiments on public benchmarks, where QuiZSF outperforms state-of-the-art baselines. Using Non-LLM TSPMs, QuiZSF achieves top performance in \textbf{75\%} of ZSF settings; with LLM-based TSPMs, it ranks first in \textbf{87.5\%} of settings, while maintaining high efficiency in memory usage and inference speed.
\end{itemize}

\begin{figure*}[ht]
    \centering
    \resizebox{0.95\textwidth}{0.35\textheight}{ 
        \includegraphics{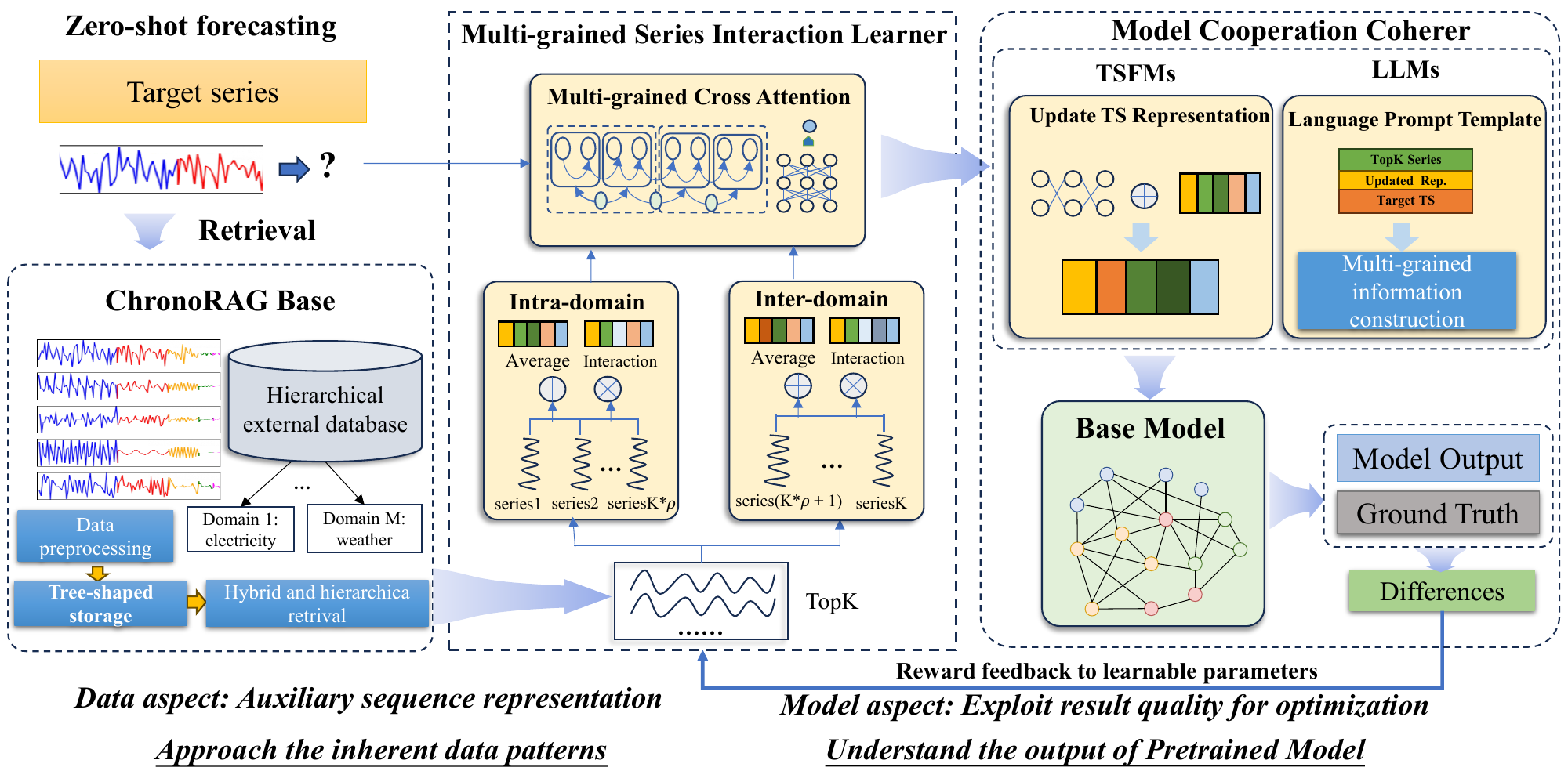}
    }
    \caption{Overview of QuiZSF.}
    \label{fig:QuiZSF}
\end{figure*}

\section{Related Work}

\subsection{TSPMs for Zero-shot Forecasting}
Recent advancements in TSPMs for ZSF have garnered significant attention. These models can be broadly categorized into two types. The first is \textbf{pre-trained models} designed specifically for TSF. These models learn generalizable temporal representations from large-scale TS datasets.Representative methods include Moment~\cite{goswami2024moment}, TTM~\cite{ekambaram2024tiny}, TimesFM~\cite{das2024decoder}, Moirai~\cite{woo2024unified}, and Lag-LLaMA~\cite{rasul2023lag}. 
The second type leverages \textbf{pre-trained large language models (LLMs)} by framing time-series forecasting as a form of cross-domain transfer learning. These models, including LLMTime~\cite{gruver2024large}, Time-LLM~\cite{jin2023time}, and GPT4TS~\cite{zhou2023one}, transform numerical sequences into natural language prompts, enabling ZSF.

Despite promising results, these models rely heavily on large-scale pre-training or fine-tuning across diverse datasets. They often lack scalability and the ability to incorporate real-time, open-world knowledge. As illustrated in Figure~\ref{fig:motivation}, time series from similar domains frequently share common temporal patterns, which can serve as valuable references in low-resource settings. With the advancement of retrieval techniques, integrating external sequence-level knowledge into pre-trained models enables dynamic forecasting, mitigates hallucinations, and constrains forecasting within realistic boundaries. Enhancing TSPMs with retrieval capabilities has therefore emerged as a promising direction for improving zero-shot time-series forecasting.
\subsection{Retrieval Augmented Generation}
Retrieval-Augmented Generation (RAG) combines models with information retrieval to enhance model performance by leveraging external knowledge. Early work, such as REALM~\cite{guu2020retrieval}, introduces retrieval-augmented models by incorporating unsupervised retrieval modules, allowing models to access relevant passages during generation. RAG~\cite{lewis2020retrieval} further advances this by combining retrieval and generation in an end-to-end framework, improving tasks like open-domain question answering. Subsequent approaches, including T5+RAG~\cite{lewis2020retrieval} and Fusion-in-Decoder (FiD)~\cite{izacard2020leveraging}, enhance the fusion of retrieved information, achieving better coherence and relevance. 
Applying RAG to time series forecasting offers promising benefits. Recent studies~\cite{jing2022retrieval, yeh2023efficient, tire2024retrieval} have introduced retrieval mechanisms into forecasting. However, they often overlook the cost and scalability of external databases in large-scale settings, leading to inefficiencies. While TimeRAF~\cite{zhang2024timeraf} explores zero-shot forecasting with retrieval, it lacks a clear mechanism for modeling interactions between pre-trained models and retrieved sequences. 
These pioneering studies highlight the potential value of combining pre-trained models with retrieval processes.

\section{Preliminaries and Problem Definition}
We focus on zero-shot forecasting (ZSF) with Time Series Pre-trained Models (TSPMs), which can be categorized into  Non-LLM based (processing numerical inputs) and LLM based (operating on language inputs). We conduct separate ZSF formulations to accommodate their distinct modalities.


\subsection{Retrieval-Augmented Non-LLM based TSPMs}
Let \(\mathcal{G} = \{\mathcal{G}_1, \dots, \mathcal{G}_K\}\) denote multiple distinct TS domains, where each domain \(\mathcal{G}_k\) contains series \(\mathcal{G}_k = \{\mathbf{X}^k \mid x^k_1, \dots, x^k_T\}\). These series are compiled into a dynamic auxiliary database \(\mathcal{D}\).

Given a target sequence \(\mathbf{X}^T\), we retrieve the top-\(K\) relevant auxiliary series \(\mathbb{X}^R = \{\mathbf{X}^R_1, \dots, \mathcal{X}^R_K\} \subset \mathcal{D}\). Since Non-LLM models do not support textual prompts, the target and retrieved series are fused into a new representation, $\widehat{\mathbf{X}}^T = \mathcal{F}_{\text{N}}^*(\mathbf{X}^T, \mathbb{X}^R), \quad \widehat{\mathbf{Y}}^T = \mathcal{M}_1^*(\widehat{\mathbf{X}}^T)$, where \(\mathcal{M}_1\) is the modified learning framework with Non-LLM based TSPM.

\subsection{Retrieval-Augmented LLM based TSPMs}
In contrast, for LLM based TSPMs, we first process the target sequence and retrieved auxiliary series to produce both a structured input and a corresponding textual prompt:
\begin{equation}
\widehat{\mathbf{X}}^T, \mathcal{P} = \mathcal{F}_{\text{L}}^*(\mathbf{X}^T, \mathbb{X}^R), \quad \widehat{\mathbf{Y}}^T = \mathcal{M}_2^*(\widehat{\mathbf{X}}^T \mid \mathcal{P}),
\end{equation}
where \(\mathcal{M}_2\) is the modified learning framework and \(\mathcal{P}\) is the generated prompt fusing the retrieved knowledge.

\section{Methodology}
\label{sec:method}
\subsection{Framework overview}

Our zero-shot learning framework consists of three major components, as illustrated in Figure~\ref{fig:QuiZSF}. (i)  a \textbf{ChronoRAG Base} with efficient hierarchical retrieval, (ii)  a \textbf{Multi-grained Series Interaction Learner} captures series-level interactions between the target and retrieved series, offering valuable information for zero-shot forecasting, (iii)  a \textbf{Model Cooperation Coherer} enables modality-aligned integration between learned representations and both Non-LLM based and LLM based TSPMs. 

\subsection{ChronoRAG Base}
ChronoRAG Base (CRB) comprises 27 time series datasets categorized into seven domains: Web, Energy, Health, IoT, Nature, Transport, and Environment. It primarily integrates five open data sources: UTSD~\cite{liutimer}, TSER Archive~\cite{tan2021time}, Monash~\cite{godahewa2021monash}, TDBrain~\cite{van2022two}, and UCR Time Series Archive~\cite{dau2019ucr}. These datasets exhibit diverse sampling frequencies, ranging from macro-level intervals (e.g., daily) to finer granularities (e.g., hourly or minute-level). Notably, some datasets demonstrate exceptionally high sampling rates, such as the TDBrain dataset, which operates at a millisecond-level frequency. Based on these datasets, we constructed three versions of the database: CRB-Small, CRB-Medium, and CRB-Large, containing 34M, 48M, and 143M time points, respectively. Detailed information about CRB-Large can be found in Table~\ref{tab:CRB-Large}. Each database covers all seven domains, and the smaller versions are subsets of the larger ones. There are three important  database designs, include a well-designed data protocol for ChronRAG for processing high-quality datasets, a  Hierarchical Series Tree for efficient series-level storage as well as a Hybrid and Hierarchical Time-series Retrieval for efficient retrieval.


\begin{table*}[]
\centering
\setlength{\tabcolsep}{2pt}
\caption{CRB-Large Detailed Descriptions: Domain indicates the field to which the dataset belongs. Datasets refer to the specific datasets included. Time Series represents the number of time series contained in the dataset after processing. Frequency denotes the sampling interval of time points, where “-” indicates either the absence of timestamps or irregular intervals. Time Points represents the total number of time points in the dataset. Source specifies the original paper or resource from which the dataset is obtained.} 
\label{tab:CRB-Large}
\resizebox{0.95\textwidth}{!}{
\begin{tabular}{c|c|c|c|c|c}
\hline
Domain      & Datasets                                                 & Time Series & Frequency & Time Points & Source                  \\ \hline
Web         & kaggle\_web\_traffic\_dataset\_without\_missing\_values  & 141444      & Daily     & 72419328    & Monash~\cite{godahewa2021monash}                  \\ \hline
\multirow{3}{*}{Energy}      & wind\_4\_seconds\_dataset                                & 1           & 4 Sec     & 512         & Monash~\cite{godahewa2021monash}                  \\ \cline{3-6} 
            & australian\_electricity\_demand\_dataset                 & 5           & 30 Min    & 2560        & Monash~\cite{godahewa2021monash}                  \\ \cline{3-6} 
            & london\_smart\_meters\_dataset\_without\_missing\_values & 5556        & Hourly    & 2844672     & Monash~\cite{godahewa2021monash}                  \\ \hline
\multirow{9}{*}{Health}       & SelfRegulationSCP1                                       & 3366        & 0.004 Sec & 1723392     & UCR Time Series Archive~\cite{dau2019ucr} \\ \cline{3-6} 
            & MotorImagery                                             & 24192       & 0.001 Sec & 12386304    & UCR Time Series Archive~\cite{dau2019ucr} \\ \cline{3-6} 
            & PigCVP                                                   & 312         & -         & 159744      & UCR Time Series Archive~\cite{dau2019ucr} \\ \cline{3-6} 
            & PigArtPressure                                           & 312         & -         & 159744      & UCR Time Series Archive~\cite{dau2019ucr} \\ \cline{3-6} 
            & SelfRegulationSCP2                                       & 2660        & 0.004 Sec & 1361920     & UCR Time Series Archive~\cite{dau2019ucr} \\ \cline{3-6} 
            & AtrialFibrillation                                       & 60          & 0.008 Sec & 30720       & UCR Time Series Archive~\cite{dau2019ucr} \\ \cline{3-6} 
            & IEEEPPG                                                  & 15480       & 0.008 Sec & 7925760     & TSER archive~\cite{tan2021time}            \\ \cline{3-6} 
            & BIDMC32HR                                                & 12278       & -         & 6286336     & TSER archive~\cite{tan2021time}            \\ \cline{3-6} 
            & TDBrain                                                 & 28644       & 0.002 Sec & 14665728    & TDBrain~\cite{van2022two}                 \\ \hline
IoT         & baian                                                    & 918         & 0.02 Sec  & 470016      & UTSD ~\cite{liutimer}                    \\ \hline
\multirow{9}{*}{Nature}       & StarLightCurves                                          & 9236        & -         & 4728832     & UCR Time Series Archive~\cite{dau2019ucr} \\ \cline{3-6} 
            & Phoneme                                                  & 2110        & -         & 1080320     & UCR Time Series Archive~\cite{dau2019ucr} \\ \cline{3-6} 
            & EigenWorms                                               & 1554        & -         & 795648      & UCR Time Series Archive~\cite{dau2019ucr} \\ \cline{3-6} 
            & Worms                                                    & 258         & 0.033 Sec & 132096      & UCR Time Series Archive~\cite{dau2019ucr} \\ \cline{3-6} 
            & us\_births\_dataset                                      & 1           & Daily     & 512         & Monash~\cite{godahewa2021monash}                  \\ \cline{3-6} 
            & kdd\_cup\_2018\_dataset\_without\_missing\_values        & 270         & Hourly    & 138240      & Monash~\cite{godahewa2021monash}                  \\ \cline{3-6} 
            & temperature\_rain\_dataset\_without\_missing\_values     & 32072       & Daily     & 16420864    & Monash~\cite{godahewa2021monash}                  \\ \cline{3-6} 
            & Sunspot\_dataset\_without\_missing\_values               & 1           & Daily     & 512         & Monash~\cite{godahewa2021monash}                  \\ \cline{3-6} 
            & saugeenday\_dataset                                      & 1           & Daily     & 512         & Monash~\cite{godahewa2021monash}                  \\ \hline

Transport   & pedestrian\_counts\_dataset                              & 66          & Hourly    & 33792       & Monash~\cite{godahewa2021monash}                  \\ \hline
\multirow{3}{*}{Environment} & AustraliaRainfall                                        & 3           & Hourly    & 1536        & TSER archive~\cite{tan2021time}            \\ \cline{3-6} 
            & BenzeneConcentration                                     & 8           & Hourly    & 4096        & TSER archive~\cite{tan2021time}            \\ \cline{3-6} 
            & BeijingPM25Quality                                       & 9           & Hourly    & 4608        & TSER archive~\cite{tan2021time}            \\ \hline
\end{tabular}
}
\end{table*}

\subsubsection{Data protocol for ChronoRAG.}    To ensure consistency and scalability, we design a unified data protocol including sliding-window segmentation, linear interpolation for missing values, channel-independent processing for dimensional alignment, and standardized metadata. All sequences are stored in the ARROW format~\cite{peltenburg2019fletcher} for efficient access. Further Details are provided in Appendix~\ref{app:data_protocol}.

\subsubsection{Hierarchical Series Tree}    It supports efficient indexing and retrieval in CRB by partitioning the database into domain-based groups and using \(k\)-means clustering. This hierarchical structure accelerates approximate nearest-neighbor search and supports dynamic updates, ensuring the index evolves incrementally over time.  Details are provided in Appendix~\ref{app:tree}.
\subsubsection{Hybrid and Hierarchical Time-series Retrieval}
\begin{figure*}[t]
    \centering
    \includegraphics[width=0.85\textwidth]{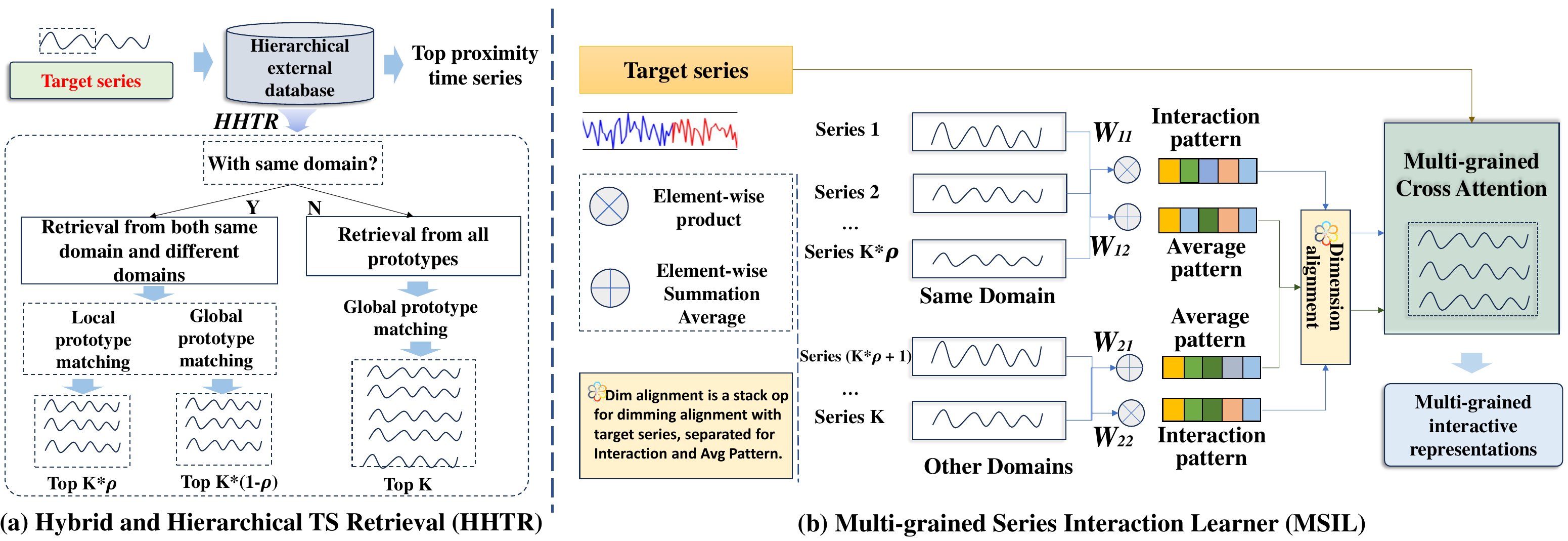}  
\caption{(a) HHTR: Domain-aware and global retrieval via a hierarchical index. (b) MSIL: Interaction and average patterns extracted from retrieved sequences are fused with the target via cross-attention.}
    \label{fig:HHTR_MSIL}
\end{figure*}

To support both high-accuracy and scalable search, we propose a hybrid and hierarchical time-series  retrieval (HHTR) strategy. It combines local domain-specific matching and global prototype comparison, leveraging the hierarchical index structure built in CRB. The strategy is shown in Figure~\ref{fig:HHTR_MSIL} (a).



\textbf{Top-K Series Retrieval.}
Given a query time series \(\mathbf{X}_T\), the objective is to retrieve its most relevant neighbors from the prototype database. This is a classical nearest-neighbor retrieval task, tailored to the time-series domain through hybrid matching strategies. When the domain of \(\mathbf{X}_T\) is known and exists in CRB, we apply a combination of local and global prototype matching. The final Top-\(K\) set is computed as:
\begin{equation}
\text{Top}\;\;K = \rho \cdot \text{Top}\;\;K_{\text{local}} + (1 - \rho) \cdot \text{Top}\;\;K_{\text{global}},
\end{equation}
where \(\rho \in [0, 1]\) controls the balance between local and global contributions. When the domain is not identified or not present in CRB, retrieval is performed across all stored cluster prototypes,
\begin{equation}
\text{Top}\;\;K = \text{Top}\;\; K_{\text{global}},
\end{equation}
Compared with existing methods, the proposed hybrid design exhibits superior performance in retrieval accuracy while maintaining higher computational efficiency.


\textbf{Distance Metric Design.} To quantify similarity between the query \(\mathbf{X}_T\) and a candidate sequence \(\mathbf{X}_i\), we define a compound similarity score:
\begin{equation}
\text{Sim}(\mathbf{X}_T, \mathbf{X}_i) = \cos(\mathbf{X}_T, \mathbf{X}_i) + \frac{1}{\text{dist}(\mathbf{X}_T, \mathbf{X}_i)},
\end{equation}
where \(\cos(\cdot, \cdot)\) denotes cosine similarity and \(\text{dist}(\cdot, \cdot)\) is the Euclidean distance. This dual metric emphasizes trend alignment through cosine similarity while capturing geometric proximity via Euclidean distance.

\subsection{Multi-grained Series Interaction Learner}
The core challenge after retrieving relevant series lies in effectively integrating them with the target series to enhance forecasting performance. Simply concatenating or averaging the retrieved series can lead to suboptimal results due to heterogeneity across domains. To this end, we propose the \textbf{Multi-grained Series Interaction Learner (MSIL)}, which is designed to extract robust and informative representations by modeling fine-grained interactions and global trends simultaneously.


MSIL is motivated by three considerations: (1) retrieved series may come from different domains with distinct dynamics; (2) fine-grained dependencies across series are often crucial for forecasting; and (3) domain-specific context and global knowledge should be fused in a unified representation. 
As illustrated in Figure~\ref{fig:HHTR_MSIL} (b), MSIL achieves this by computing two representative patterns, namely an interaction pattern and an average pattern, which are then fused with the target sequence through a cross-attention mechanism.

Given a target time series \(\mathbf{T} \in \mathbb{R}^{N \times D}\) with \(N\) time steps and \(D\) channels, and a set of retrieved series \(\{\mathbf{S}_1, \dots, \mathbf{S}_n\}\), we first divide the retrieved set into same-domain and cross-domain subsets:
\begin{equation}
\mathbf{S}_i \in 
\begin{cases} 
\mathbb{S}_{\text{same}} & \text{if } \text{domain}(\mathbf{S}_i) = \text{domain}(\mathbf{T}), \\
\mathbb{S}_{\text{cross}} & \text{if } \text{domain}(\mathbf{S}_i) \neq \text{domain}(\mathbf{T})
\end{cases}
\quad \forall i \in n
\end{equation}


To ensure numerical consistency, we normalize both target and retrieved sequences using the scaler module associated with their respective base models:
\begin{align}
    \mathbf{T}^{\text{norm}}, \text{loc}_T, \text{scale}_T &= \text{scaler}(\mathbf{T}), \\
    \mathbf{S}_i^{\text{norm}}, \text{loc}_{S_i}, \text{scale}_{S_i} &= \text{scaler}(\mathbf{S}_i), 
    \quad \forall i = 1, 2, \ldots, K
\end{align}
with
\begin{equation}
    \mathbf{T}^{\text{norm}} = \frac{\mathbf{T} - \text{loc}_T}{\text{scale}_T}, \quad 
    \mathbf{S}_i^{\text{norm}} = \frac{\mathbf{S}_i - \text{loc}_{S_i}}{\text{scale}_{S_i}}
\end{equation}

Based on the normalized retrieved sequences, MSIL computes two complementary patterns: \textbf{1) Interaction Pattern} (\(\mathbf{P}_{\text{int}}\)) captures fine-grained dependencies via element-wise product, followed by a non-linear projection. \textbf{2) Average Pattern} (\(\mathbf{P}_{\text{avg}}\)) encodes global trends through mean pooling and transformation. These are defined as:
\begin{equation}
\begin{aligned}
    \mathbf{P}_{\text{int}} &= \text{MLP}_1\left( \frac{\prod_{i=1}^{n} \mathbf{S}_i^{\text{norm}}}{\left\|\prod_{i=1}^{n} \mathbf{S}_i^{\text{norm}}\right\|} \right), &\quad
    \mathbf{P}_{\text{avg}} &= \text{MLP}_2\left( \frac{1}{n} \sum_{i=1}^{n} \mathbf{S}_i^{\text{norm}} \right)
\end{aligned}
\end{equation}


To fuse these patterns with the target sequence, we use a multi-grained cross-attention mechanism, where the target serves as query and the patterns as key/value:
\begin{align}
    \mathbf{Q} = \mathbf{W}_q \mathbf{T}^{\text{norm}}, \quad
    \mathbf{K} = \mathbf{W}_k \mathbf{P}_{\text{avg}}, \quad
    \mathbf{V} = \mathbf{W}_v \mathbf{P}_{\text{int}},
\end{align}
\begin{equation}
    \mathbf{R}_{\text{fused}} = \text{Softmax}\left(
        \frac{\mathbf{Q} \cdot \mathbf{K}^T}{\sqrt{d}}
    \right) \cdot \mathbf{V}
\end{equation}

The resulting representation \(\mathbf{R}_{\text{fused}}\) combines domain-aware alignment and trend-aware feature fusion. 
MSIL enables rich interaction modeling between the target and retrieved series, and its multi-granular design improves generalization, especially in zero-shot forecasting scenarios.

\begin{table*}[!t]
\centering
\caption{Long sequence forecasting results. The experimental setup follows TTM~\cite{ekambaram2024tiny}. Best results are in bold; second best are underlined. Full-shot results are obtained from the Moirai~\cite{woo2024unified} where the authors
draw similar comparison.}
\label{tab:zero_shot}

\resizebox{0.9\linewidth}{!}{  

\begin{tabular}{ccccccccccccc}

\toprule[1.2pt]
                         &     & \multicolumn{4}{c}{Zero-shot forecasting}                                                                                                                                                                                                      & \multicolumn{7}{c}{Full-shot forecasting}                                                                                                                                                                                                                                                                                                                                                                                            \\ 
\cmidrule(lr){3-6} \cmidrule(l){7-13} 
\multicolumn{2}{c}{}           & \textbf{\begin{tabular}[c]{@{}c@{}}QuiZSF$_{T}$\\ (Ours)\end{tabular}} & \begin{tabular}[c]{@{}c@{}}TTM$_B$\\ (2024)\end{tabular} & \begin{tabular}[c]{@{}c@{}}Moirai$_B$\\ (2024)\end{tabular} & \begin{tabular}[c]{@{}c@{}}TimesFM\\ (2024)\end{tabular} & \begin{tabular}[c]{@{}c@{}}iTransformer\\ (2024)\end{tabular} & \begin{tabular}[c]{@{}c@{}}Crossformer\\ (2023)\end{tabular} & \begin{tabular}[c]{@{}c@{}}DLinear\\ (2023)\end{tabular} & \begin{tabular}[c]{@{}c@{}}TimesNet\\ (2023)\end{tabular} & \begin{tabular}[c]{@{}c@{}}PatchTST\\ (2023)\end{tabular} & \begin{tabular}[c]{@{}c@{}}TiDE\\ (2023)\end{tabular} & \begin{tabular}[c]{@{}c@{}}FEDformer\\ (2022)\end{tabular} \\ 
\toprule[1.1pt]
\multirow{5}{*}{ETTh1}   & 96  & \textbf{0.361}                                                   & \underline{0.364}                                           & 0.384                                                    & 0.421                                                    & 0.386                                                         & 0.423                                                        & 0.386                                                    & 0.384                                                     & 0.414                                                     & 0.479                                                 & 0.376                                                      \\
                         & 192 & \textbf{0.384}                                                   & \underline{0.388}                                           & 0.425                                                    & 0.472                                                    & 0.441                                                         & 0.471                                                        & 0.437                                                    & 0.436                                                     & 0.46                                                      & 0.525                                                 & 0.420                                                      \\
                         & 336 & \textbf{0.398}                                                   & \underline{0.402}                                           & 0.456                                                    & 0.51                                                     & 0.487                                                         & 0.570                                                        & 0.481                                                    & 0.491                                                     & 0.501                                                     & 0.565                                                 & 0.459                                                      \\
                         & 720 & \textbf{0.468}                                                   & \underline{0.471}                                           & 0.470                                                    & 0.514                                                    & 0.503                                                         & 0.653                                                        & 0.519                                                    & 0.521                                                     & 0.500                                                     & 0.594                                                 & 0.506                                                      \\ \cline{2-13} 
                         & Avg & \textbf{0.403}                                                   & \underline{0.406}                                           & 0.434                                                    & 0.479                                                    & 0.454                                                         & 0.529                                                        & 0.456                                                    & 0.458                                                     & 0.469                                                     & 0.541                                                 & 0.440                                                      \\ 
\toprule[0.8pt]
\multirow{5}{*}{ETTh2}   & 96  & \textbf{0.276}                                                   & 0.279                                                 & \underline{0.277}                                              & 0.326                                                    & 0.297                                                         & 0.745                                                        & 0.333                                                    & 0.34                                                      & 0.302                                                     & 0.4                                                   & 0.358                                                      \\
                         & 192 & \textbf{0.334}                                                   & \textbf{0.334}                                        & \underline{0.34}                                               & 0.400                                                    & 0.380                                                         & 0.877                                                        & 0.477                                                    & 0.402                                                     & 0.388                                                     & 0.528                                                 & 0.429                                                      \\
                         & 336 & \textbf{0.364}                                                      & \underline{0.366}                                        & 0.371                                                    & 0.434                                                    & 0.428                                                         & 1.043                                                        & 0.594                                                    & 0.452                                                     & 0.426                                                     & 0.643                                                 & 0.496                                                      \\
                         & 720 & \underline{0.407}                                                      & 0.408                                                 & \textbf{0.394}                                           & 0.451                                                    & 0.427                                                         & 1.104                                                        & 0.831                                                    & 0.462                                                     & 0.431                                                     & 0.874                                                 & 0.463                                                      \\ \cline{2-13} 
                         & Avg & \textbf{0.345}                                                   & 0.347                                           & \underline{0.346}                                              & 0.403                                                    & 0.383                                                         & 0.942                                                        & 0.559                                                    & 0.414                                                     & 0.388                                                     & 0.611                                                 & 0.437                                                      \\ 
\toprule[0.8pt]
\multirow{5}{*}{ETTm1}   & 96  & 0.369                                                            & 0.359                                                & 0.335                                                    & 0.357                                                    & \underline{0.334}                                                   & 0.404                                                        & 0.345                                                    & 0.338                                                     & \textbf{0.329}                                            & 0.364                                                 & 0.379                                                      \\
                         & 192 & 0.377                                                            & 0.376                                                 & \underline{0.367}                                              & 0.411                                                    & \textbf{0.337}                                                & 0.450                                                        & 0.38                                                     & 0.374                                                     & \underline{0.367}                                               & 0.398                                                 & 0.426                                                      \\
                         & 336 & \textbf{0.397}                                                   & 0.407                                                 & \underline{0.398}                                              & 0.442                                                    & 0.426                                                         & 0.532                                                        & 0.413                                                    & 0.41                                                      & 0.409                                                     & 0.428                                                 & 0.445                                                      \\
                         & 720 & \underline{0.441}                                                      & 0.446                                                 & \textbf{0.434}                                           & 0.507                                                    & 0.491                                                         & 0.666                                                        & 0.474                                                    & 0.478                                                     & 0.481                                                     & 0.487                                                 & 0.543                                                      \\ \cline{2-13} 
                         & Avg & \underline{0.395}                                                      & 0.397                                                 & \textbf{0.383}                                           & 0.429                                                    & 0.397                                                         & 0.513                                                        & 0.403                                                    & 0.400                                                     & 0.397                                               & 0.419                                                 & 0.448                                                      \\ 
\toprule[0.8pt]
\multirow{5}{*}{ETTm2}   & 96  & \underline{0.176}                                                      & 0.178                                                 & 0.195                                                    & 0.205                                                    & 0.18                                                          & 0.287                                                        & 0.193                                                    & 0.187                                                     & \textbf{0.175}                                            & 0.207                                                 & 0.203                                                      \\
                         & 192 & \textbf{0.238}                                                   & \textbf{0.238}                                        & 0.247                                                    & 0.293                                                    & 0.25                                                          & 0.414                                                        & 0.284                                                    & 0.249                                                     & \underline{0.241}                                               & 0.290                                                 & 0.269                                                      \\
                         & 336 & \textbf{0.292}                                                   & 0.300                                                   & \underline{0.293}                                              & 0.366                                                    & 0.311                                                         & 0.597                                                        & 0.369                                                    & 0.321                                                     & 0.305                                                     & 0.377                                                 & 0.325                                                      \\
                         & 720 & \underline{0.390}                                                      & 0.41                                                  & \textbf{0.365}                                           & 0.472                                                    & 0.412                                                         & 1.730                                                        & 0.554                                                    & 0.408                                                     & 0.402                                                     & 0.558                                                 & 0.421                                                      \\ \cline{2-13} 
                         & Avg & \textbf{0.274}                                                   & 0.282                                                 & \underline{0.275}                                              & 0.334                                                    & 0.288                                                         & 0.757                                                        & 0.350                                                    & 0.291                                                     & 0.281                                                     & 0.358                                                 & 0.305                                                      \\ 
\toprule[0.8pt]
\multirow{5}{*}{Weather} & 96  & \textbf{0.153}                                                   & \underline{0.158}                                           & 0.167                                                    & -                                                        & 0.174                                                         & \underline{0.158}                                                  & 0.196                                                    & 0.172                                                     & 0.177                                                     & 0.202                                                 & \underline{0.217}                                                \\
                         & 192 & \textbf{0.194}                                                   & \underline{0.206}                                           & 0.209                                                    & -                                                        & 0.221                                                         & 0.206                                                        & 0.237                                                    & 0.219                                                     & 0.225                                                     & 0.242                                                 & 0.276                                                      \\
                         & 336 & \textbf{0.251}                                                   & 0.260                                           & \underline{0.256}                                              & -                                                        & 0.278                                                         & 0.272                                                        & 0.283                                                    & 0.280                                                     & 0.278                                                     & 0.287                                                 & 0.339                                                      \\
                         & 720 & \textbf{0.324}                                                   & 0.330                                                 & \underline{0.325}                                              & -                                                        & 0.358                                                         & 0.398                                                        & 0.345                                                    & 0.365                                                     & 0.354                                                     & 0.351                                                 & 0.403                                                      \\ \cline{2-13} 
                         & Avg & \textbf{0.231}                                                   & \underline{0.239}                                           & \underline{0.239}                                                    & -                                                        & 0.258                                                         & 0.259                                                        & 0.265                                                    & 0.259                                                     & 0.259                                                     & 0.271                                                 & 0.309                                                      \\ 
\toprule[1.2pt]
\end{tabular}
}
\end{table*}

\subsection{Model Cooperation Coherer}

In retrieval-augmented zero-shot time-series forecasting, it is crucial to effectively connect retrieved knowledge with diverse TSPMs. These models fall into two categories: Non-LLM-based (numerical input) and LLM-based (textual input), each requiring tailored integration strategies. To fully leverage the representations produced by MSIL, we design a unified cooperation mechanism with two branches: one for numerical models and one for language models, both supporting feedback-driven optimization.

\textbf{Numerical Coherer for Non-LLM based TSPMs.}
For numerical TS pre-trained models, we apply a residual connection to fuse the normalized target sequence \( \mathbf{T}^{\text{norm}} \) with the MSIL-fused representation \( \mathbf{R}_{\text{fused}} \), then feed it into the forecasting model:
\begin{equation}
\widehat{\mathbf{T}} = \mathcal{F}_{\text{N}}(\mathbf{R}_{\text{fused}}, \mathbf{T}^{\text{norm}})
\end{equation}
where \( \mathcal{F}_{\text{N}} \) is a residual module~\cite{he2016deep} that enhances expressiveness and mitigates gradient vanishing, and the output \(\widehat{\mathbf{T}}\) denotes the forecasted sequence.

\textbf{Language Coherer for LLM based TSPMs.}
Language models operate exclusively on text, making direct use of numeric features infeasible. To bridge this modality gap, we convert MSIL outputs (\(\mathbf{P}_{\text{int}}, \mathbf{P}_{\text{avg}}, \mathbf{T}^{\text{norm}}\)) into structured textual summaries. These summaries, along with an instruction-style prompt, guide the language model in generating forecasting outputs. See Appendix~\ref{app:language_coherer} and Figure~\ref{fig:prompt} for prompt construction details.

\section{Experiments}
\label{sec:experiment}
\subsection{Experiments Setups}
\label{sec:setup}
For  \textbf{evaluation datasets}, we use five public datasets: ETTh1, ETTh2, ETTm1, ETTm2, and Weather, which are widely used in prior state-of-the-art works~\cite{nie2022time, jin2023time, zhou2023one}. Standard error metric is MSE. For \textbf{CRB selection}, our ChronoRAG Base comes in three versions: CRB-Small, CRB-Medium, and CRB-Large. Given the balance between computing time and experimental results, we select CRB-Medium as the RAG Base. 
For  \textbf{model comparison}, we evaluate against thirteen state-of-the-art forecasting methods, which can be classified into the following categories: (a) \textbf{Non-LLM based TSPMs:} TTM~\cite{ekambaram2024tiny}, Moirai~\cite{woo2024unified}, and TimesFM~\cite{das2024decoder}; (b) \textbf{LLM based TSPMs:} TimeLLM~\cite{jin2023time}, LLMTime~\cite{gruver2024large}, and GPT4TS~\cite{zhou2023one}; (c) \textbf{Other architectures:} iTransformer~\cite{liu2023itransformer}, Crossformer~\cite{zhang2023crossformer}, DLinear~\cite{zeng2023transformers}, TimesNet~\cite{wu2022timesnet}, TiDE~\cite{das2023long}, PatchTST~\cite{nie2022time}, and FEDformer~\cite{zhou2022fedformer}.
\subsection{Implementation Details}


In the time series domain, \textbf{zero-shot time-series forecasting} refers to evaluating models on unseen datasets without direct supervision or fine-tuning. Given the two types of TSPMs, current ZSF setups are categorized accordingly: the \textbf{multi-source generalization setup} for Non-LLM based models, and the \textbf{single-source transfer setup} for LLM based models. To ensure comprehensive evaluation, we adopt both setups in our experiments, resulting in \textbf{QuiZSF$_T$} and \textbf{QuiZSF$_L$}, respectively.

\textbf{Multi-source generalization zero-shot setup (for Non-LLM based TSPMs)}
This setup trains on diverse source datasets and enables zero-shot forecasting by directly applying the model to unseen target datasets.
This approach is widely used in Non-LLM based TSPMs like TTM~\cite{ekambaram2024tiny}, TimesFM~\cite{das2024decoder}, and Moirai~\cite{woo2024unified}. We use this setup for \textbf{QuiZSF$_T$}, with TTM-Base~\cite{ekambaram2024tiny} as the base model. 
For training, we use a comprehensive dataset covering 38.7 million time points, curated from multiple public benchmarks to ensure diversity.
Evaluation is conducted on the held-out ETT and Weather datasets, ensuring strict zero-shot conditions without data leakage. 
Details of the training datasets can be found in Appendix~\ref{app:train_datasets}.

\begin{table*}[!t]
\centering
\caption{Zero-shot forecasting results under the single-source transfer setup. Following the setting of TimeLLM~\cite{jin2023time}, results are averaged over prediction lengths \{96, 192, 336, 720\}. Best scores are in bold, second best are underlined.}
\label{tab:A->B}

\resizebox{0.9\linewidth}{!}{
\begin{tabular}{c|c|c|c|c|c|c|c|c}
\hline
                         & \textbf{QuiZSF\textsubscript{L}} & Time-LLM & LLMTime & GPT4TS & DLinear & PatchTST & TimesNet & Autoformer \\ \hline
ETTh1 $\rightarrow$ ETTh2 & \textbf{0.352} & \underline{0.356} & 0.992   & 0.406  & 0.493   & 0.380    & 0.421    & 0.582      \\ \hline
ETTh1 $\rightarrow$ ETTm2 & \textbf{0.272} & \underline{0.277} & 1.867   & 0.325  & 0.415   & 0.314    & 0.327    & 0.457      \\ \hline
ETTh2 $\rightarrow$ ETTh1 & \underline{0.535} & \textbf{0.521} & 1.961   & 0.757  & 0.703   & 0.545    & 0.865    & 0.757      \\ \hline
ETTh2 $\rightarrow$ ETTm2 & \textbf{0.269} & \underline{0.271} & 1.867   & 0.335  & 0.328   & 0.325    & 0.342    & 0.366      \\ \hline
ETTm1 $\rightarrow$ ETTh2 & \textbf{0.382} & \underline{0.394} & 0.992   & 0.433  & 0.464   & 0.439    & 0.457    & 0.470      \\ \hline
ETTm1 $\rightarrow$ ETTm2 & \textbf{0.281} & 0.296             & 1.867   & 0.313  & 0.335   & \underline{0.291} & 0.322    & 0.469      \\ \hline
ETTm2 $\rightarrow$ ETTh2 & \textbf{0.351} & \underline{0.354} & 0.992   & 0.435  & 0.455   & 0.409    & 0.435    & 0.423      \\ \hline
ETTm2 $\rightarrow$ ETTm1 & \textbf{0.414} & \underline{0.418} & 1.993   & 0.769  & 0.649   & 0.568    & 0.769    & 0.755      \\ \hline
\end{tabular}
}
\end{table*}

\begin{table}[]
\centering
\small
\caption{Ablation studies on zero-shot settings}
  \begin{tabular}{ccc}
  \hline
               & ETTm1 → ETTh2  & ETTm1 → ETTm2  \\ \hline
  QuiZSF$_{L}$-w/o-RAG     & 0.394          & 0.296          \\ 
  QuiZSF$_{L}$-w/o-MSIL    & 0.387          & 0.286          \\ 
  QuiZSF$_{L}$-w/o-Coherer & 0.388          & 0.289          \\ \hline
  QuiZSF$_{L}$         & \textbf{0.382} & \textbf{0.281} \\ \hline
  \end{tabular}

\label{tab:ablation}
\end{table}

\textbf{Single-source transfer setup (for LLM based TSPMs)}
This setup trains a model on a single source dataset (e.g., ETTh1) and evaluates it on an unseen target dataset (e.g., ETTm2), highlighting the model's ability to transfer knowledge across domains. Commonly used in LLM based TSPMs~\cite{jin2023time, zhou2023one}, this setup involves fine-tuning or prompting LLMs on a specific dataset and then applying them to novel inputs from a different domain. We use this setup for \textbf{QuiZSF$_L$}, with TimeLLM~\cite{jin2023time} as the base model and LLaMA-7B~\cite{touvron2023llama} as the backbone. 
The evaluation is carried out in multiple train-test splits according to the experimental setup of TimeLLM zero-shot forecasting (Table~\ref{tab:A->B}).

\subsection{Peformance Comparison}
The prediction results of \textbf{QuiZSF$_T$} and \textbf{QuiZSF$_L$} are shown in Table~\ref{tab:zero_shot} and Table~\ref{tab:A->B}, respectively. The best performance is marked in bold, and the second best is underlined.

\begin{figure*}[ht]
    \centering
    \resizebox{0.9\textwidth}{0.22\textheight}{ 
        \includegraphics{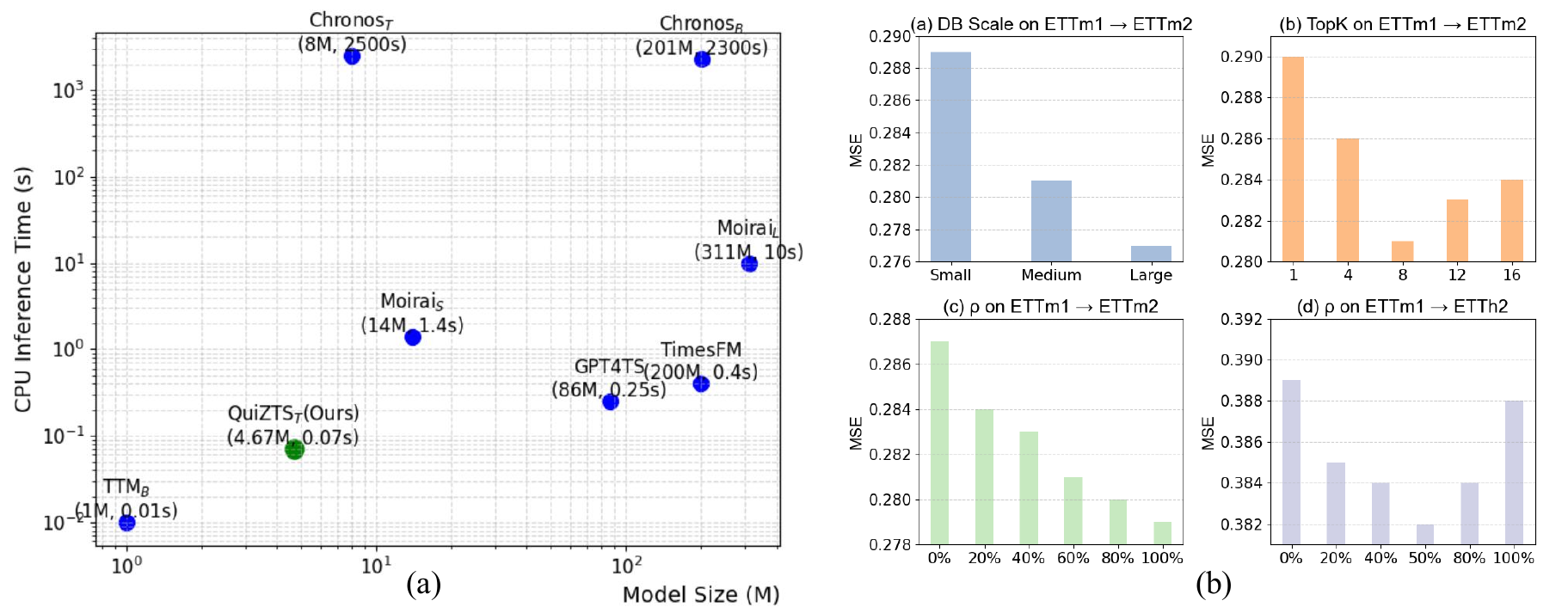}
    }
    \caption{(a) Size and time overview of QuiZTST vs. pre-trained TS benchmarks. Plot each model based on  model size and the CPU inference time per batch. (b) Hyperparameter analysis on ETTm1 -> ETTm2.}
    \label{fig:analysis}
\end{figure*}

\textbf{QuiZSF$_T$. } 
Currently, there is limited work on zero-shot forecasting. To provide a more comprehensive assessment, we compare QuiZSF$_T$ not only with zero-shot methods but also with several strong full-shot forecasting models. These comparisons fall under the category of long-sequence forecasting.
As shown in Table~\ref{tab:zero_shot}, \textbf{"zero-shot"} refers to the prediction results of various base models without any pre-training on the test datasets, while \textbf{"full-shot"} denotes the performance of benchmark models that have been fully trained on each dataset. QuiZSF$_T$, trained solely under the zero-shot setting, outperforms not only existing zero-shot baselines but also full-shot models, demonstrating strong generalization capabilities even without access to target-domain training data. QuiZSF$_T$ ranks Top1 in \textbf{75\%} of ZSF settings.
However, we observe that QuiZSF$_T$ performs particularly well on relatively coarse-grained datasets but shows limited effectiveness on short-term, minute-scale forecasting tasks (e.g., ETTm1). This may be due to the difficulty in aligning retrieved information with fine-grained fluctuations in the target sequence. Leveraging coarse-grained knowledge to guide fine-grained prediction presents an interesting direction for future research. Combining  complexity and efficiency comparison in Figure~\ref{fig:analysis} (a), we consider QuiZSF$_{T}$ to be an excellent lightweight zero-shot forecasting framework, which benefits from the retrieval-enhanced representation and  active feedback.
To further validate the generalization capability of QuiZSF$_{T}$, we also evaluate its performance when the domain of the target series is not within the seven domains of CRB. For this purpose, we use the Weather dataset (Meteorological Domain) as the test set for experimentation, with results shown in Table~\ref{tab:zero_shot}. QuiZSF$_{T}$ achieves the best performance across all prediction horizons, demonstrating its strong generalization capability.


\textbf{QuiZSF$_L$.}  As shown in Table~\ref{tab:A->B}, QuiZSF$_L$, equipped with retrieved augmented series, outperforms the majority of competing methods, achieving the best results in 7 out of 8 prediction settings. 
Additionally, our approach enhances the performance based on the pre-trained model. 
With the continuous advancement of pre-trained models, QuiZSF$_{L}$ will bring about further performance improvements when adapting to new model frameworks.

\begin{figure*}[ht]
    \centering
    \resizebox{0.93\textwidth}{0.20\textheight}{ 
        \includegraphics{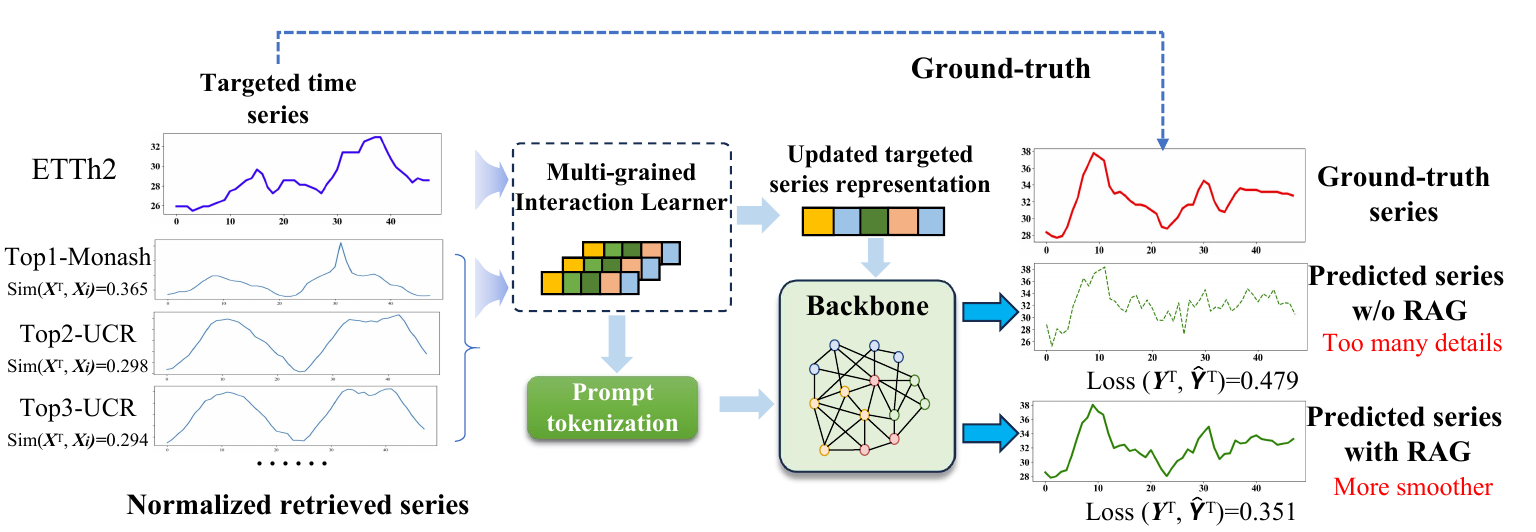}
    }
    \caption{Case studies on ETTh2 prediction.}
    \label{fig:sample}
\end{figure*}

\subsection{Ablation study}

\textbf{Ablative variants.} \textbf{1) QuiZSF$_{L}$-w/o-RAG.} We remove the auxiliary sequence retrieval and only utilize the LLM for ZSF to verify the motivation for RAG, which degenerates to LLMTime~\citep{gruver2024large}. 
\textbf{2) QuiZSF$_{L}$-w/o-MSIL.} We remove the Multi-grained Series Interaction Learner, performing only an average calculation on the retrieved time series instead of feature extraction.
\textbf{3) QuiZSF$_{L}$-w/o-Coherer.} We remove the structured prompt template and directly concatenate MSIL-extracted features with the target sequence, without adapting them into LLM-compatible textual inputs.
\textbf{Main results.} Results are presented in Table~\ref{tab:ablation}, with key findings summarized as follows:
(i) The most significant performance drop occurs when removing the retrieval module (RAG), confirming its importance in providing external contextual signals for zero-shot forecasting. Compared to the full model, this variant sees a performance decline of 3.14\%–5.34\% (\textit{line 1 vs. line 4}).
(ii) Removing MSIL and using simple averaging instead also degrades performance. 
This is because the model cannot learn multi-level representations and capture relationships between sequences, resulting in performance degradation (\textit{line 2 vs. line 4}).
(iii) Additionally, without converting numerical time series into LLM-understandable tokens using the prompt template, the LLM's performance is inferior to the full QuiZSF$_L$, with a drop of about 2\% (\textit{line 3 vs. line 4}). These results demonstrate the effectiveness of our integrated approach.

\subsection{Detailed model analysis}
\label{sec:hyper}

\textbf{Complexity analysis.} \label{sec:complexity}
We report empirical comparisons of model size and inference time in Figure~\ref{fig:analysis} (a). QuiZSF$_T$ maintains competitive efficiency while outperforming most baselines. Although it introduces minor computational overhead due to retrieval and interaction modules, the added cost is minimal. Full analysis is provided in Appendix~\ref{app:complexity}.

\textbf{Hyperparameter analysis.} Key hyperparameters include the size of the retrieval database, the number of retrieved sequences $K$, and the local-domain proportion $\rho$. As shown in Figure~\ref{fig:analysis} (b), performance is sensitive to these choices. CRB-Medium, $K=8$, and $\rho=60\%$ strike a good balance between accuracy and efficiency. Detailed experiments and discussion are provided in Appendix~\ref{app:hyper}.

\subsection{Case study}
In order to demonstrate how the retrieval sequence improves the prediction effect, we conduct an intuitive analysis of the intermediate results. Taking the target sequence of the ETTh2 dataset as an example (as shown in Figure~\ref{fig:sample}), we screen out the Top-8 sequences that are closest to the target sequence through hybrid and hierarchical time-series retrieval and marked their similarities. These sequences have similar patterns and evolution models as the target sequence. Subsequently, we update the sequence representation of the target sequence by combining the auxiliary sequence with the target sequence through MSIL and input it into the LLM after generating prompts. We visualized and compared the output of the LLM with RAG and the output w/o RAG. 

The results show that RAG can reveal the average pattern of the retrieval sequence, making the prediction results smoother and avoiding overfitting; while the output w/o RAG fluctuates more and contains more inaccurate details. This indicates that RAG effectively suppresses the time-series hallucination of the LLM. Our analysis enhances the interpretability of the model, deepens the understanding of zero-shot forecasting, and highlights the contribution of RAG in enhancing prediction.

\section{Conclusion}

We presented \textbf{QuiZSF}, a retrieval-augmented framework for zero-shot time series forecasting that unifies search and forecasting in dynamic Web environments. QuiZSF combines a tree-structured temporal database, hybrid retrieval strategies, multi-level interaction learning, and modality-aware adaptation, enabling models to retrieve auxiliary sequences and incorporate them for more accurate prediction. Experiments on public benchmarks show that QuiZSF achieves state-of-the-art performance with both Non-LLM and LLM-based TSPMs, while maintaining efficiency in memory and inference. 
Beyond forecasting, this work illustrates how retrieval-augmented AI can extend beyond text to time series, offering new perspectives for the WWW community and opening opportunities for adaptive and intelligent Web-scale systems.

\section{acknowledgements}
This paper is partially supported by the National Natural Science Foundation of China (No.62502488, No.12227901), Natural Science Foundation of Jiangsu Province (BK20240460), the grant from State Key Laboratory of Resources and Environmental Information System.

\bibliographystyle{ACM-Reference-Format}
\balance
\bibliography{reference}

\newpage
\appendix

\section{Detailed Design of ChronoRAG Base}
\label{app:crb}

\subsection{Data protocol for ChronoRAG}
\label{app:data_protocol}
Large-scale time series datasets are essential for retrieval tasks, but constructing the ChronoRAG Base (CRB) presents challenges such as inconsistent lengths, dimensionality mismatches, missing values, metadata diversity, and storage scalability. To tackle these issues, we design a unified data protocol that ensures consistent preprocessing, metadata unification, and efficient storage. All data is stored using the ARROW format~\cite{peltenburg2019fletcher}, which is optimized for deep learning frameworks and enables efficient retrieval and access.

To address the issue of \textbf{data length inconsistencies across datasets}, we adopt a sliding window approach. Let \(\mathbf{X} \in \mathbb{R}^{N \times D}\) be a multivariate time series with \(N\) time steps and \(D\) channels. The \(j\)-th channel is denoted as \(\mathbf{x}_j = [x_{1j}, x_{2j}, \cdots, x_{Nj}]\). Following the foundation model settings~\cite{ekambaram2024tiny}, we set window size \(w\) and step size \(s\) (\(w \leq N\)), and segment each channel into uniform windows:
\begin{equation}
\mathbf{X}_{kj} = [x_{(k - 1)s + 1,j}, \cdots, x_{(k - 1)s+w,j}]
\end{equation}
where \((k - 1)s + w \leq N\). This preserves local patterns and improves retrieval efficiency.

To address the issue of \textbf{varying dimensionality}, we use a channel-independent strategy, which treats each dimension separately and has been validated by PatchTST~\cite{nie2022time}, TimeLLM~\cite{jin2023time}, and TTM~\cite{ekambaram2024tiny}. For each channel \(\mathbf{x}_i\), we apply a shared function \(f\):
\begin{equation}
\mathbf{y}_i = f(\mathbf{x}_i)
\end{equation}
This simplifies database construction and fusion, while enhancing scalability and cross-domain adaptability.

To address the issue of \textbf{missing values}, which may impair data integrity and affect retrieval, we apply linear interpolation~\cite{friedman1962interpolation} to complete incomplete sequences.

To address the issue of \textbf{diverse metadata}, we define a unified metadata protocol by standardizing key attributes such as item ID, start time, end time, frequency, domain, and sequence values (see Table~\ref{tab:structure}). This ensures consistent integration across multi-source datasets.

To address the issue of \textbf{large volume and variety of sequence data}, which make efficiently storing and retrieving a significant challenge, we implement a hierarchical tree-like storage structure~\cite{berchtold1996x, jensen2004query, athanassoulis2014bf}. This enables efficient storage and indexing for large-scale datasets and seamless integration into deep learning frameworks. Details are provided in Section~\ref{app:tree}.

\begin{table}[!h]
\caption{Structural key-value instance in CRB}
\resizebox{0.46\textwidth}{!}{
\begin{tabular}{c|ccccc|c}
\hline
                     & \multicolumn{5}{c|}{Meta information}                                                                                                                                                                                      & Deterministic observation \\ \hline
\multirow{2}{*}{Key} & \multirow{2}{*}{\begin{tabular}[c]{@{}c@{}}Domain \\ Category\end{tabular}} & \multirow{2}{*}{Item\_id}                                            & \multirow{2}{*}{Start} & \multirow{2}{*}{End} & \multirow{2}{*}{Freq} & \multirow{2}{*}{Target}   \\
                     &                                                                             &                                                                      &                        &                      &                       &                           \\ \hline
Value                & Nature                                                                      & \begin{tabular}[c]{@{}c@{}}us\_births\_\\ dataset\_0\_0\end{tabular} & 20000101               & 20010527             & Daily                 & {[}9083,8006,11136,……{]}  \\ \hline
\end{tabular}
}
\label{tab:structure}
\end{table}

\subsection{Hierarchical Series Tree}

To support efficient indexing and retrieval in ChronoRAG Base (CRB), we design a hierarchical tree structure with pre-clustering. Traditional FIFO-based linear storage~\cite{pelkonen2015gorilla, jensen2017time} suffers from inefficiency when scaling to millions of time series. Linear retrieval requires one-by-one comparisons with time complexity \(T_{\text{linear}} = O(N)\), which becomes prohibitive at large scale.

To mitigate this, we construct a tree-shaped structure inspired by database indexing techniques~\cite{berchtold1996x, jensen2004query, athanassoulis2014bf}. At the top level of the structure, the database is partitioned by domain: given a dataset \(\mathcal{X} = \{X_1, \dots, X_N\}\), we separate it into \(K\) disjoint domain-based groups \(\{\mathcal{D}_k\}_{k=1}^K\), such that:
\begin{equation}
\mathcal{X} = \bigcup_{k=1}^K \mathcal{D}_k, \quad \mathcal{D}_i \cap \mathcal{D}_j = \emptyset \text{ for } i \neq j.
\end{equation}

Each domain group \(\mathcal{D}_k\) is then recursively divided using the \(k\)-means algorithm, where each cluster contains at most \(N = 256\) time series. This setting follows the clustering granularity used in prior work such as Marigold~\cite{mortensen2023marigold}, where small prototype groups (typically \(N \leq 256\)) are shown to improve retrieval quality and update flexibility. The number of clusters \(M_k\) in each domain is thus determined by the size of \(\mathcal{D}_k\), i.e., \(M_k = \lceil \frac{|\mathcal{D}_k|}{N} \rceil\).

Each cluster \(\mathcal{C}_m^{(k)}\) is formed by minimizing the standard \(k\)-means objective function:
\begin{equation}
\mathcal{L}_k = \sum_{X_i \in \mathcal{D}_k} \| X_i - C_{j(i)}^{(k)} \|^2,
\end{equation}
where \(C_{j(i)}^{(k)}\) is the centroid of the cluster to which \(X_i\) belongs. For each cluster, the prototype is selected as the sequence closest to its centroid:
\begin{equation}
X_{\text{proto}}^{(k,m)} = \arg\min_{X_i \in \mathcal{C}_m^{(k)}} \| X_i - C_m^{(k)} \|^2.
\end{equation}

While tree-based structures have a theoretical average-case complexity of \(O(\log_b N)\), this does not always hold in high-dimensional time series due to the curse of dimensionality~\cite{beyer1999nearest}. Hence, instead of relying solely on theoretical claims, we report empirical improvements in retrieval speed and memory in Section~\ref{sec:complexity}.

This hierarchical prototype-based structure supports domain-level filtering and accelerates approximate nearest-neighbor search. During retrieval, a query is first matched against cluster prototypes, and then only a small number of candidate clusters are examined in full. This greatly reduces computation compared to flat comparisons.

\begin{figure}[!h]
    \centering
    \includegraphics[width=0.45\textwidth]{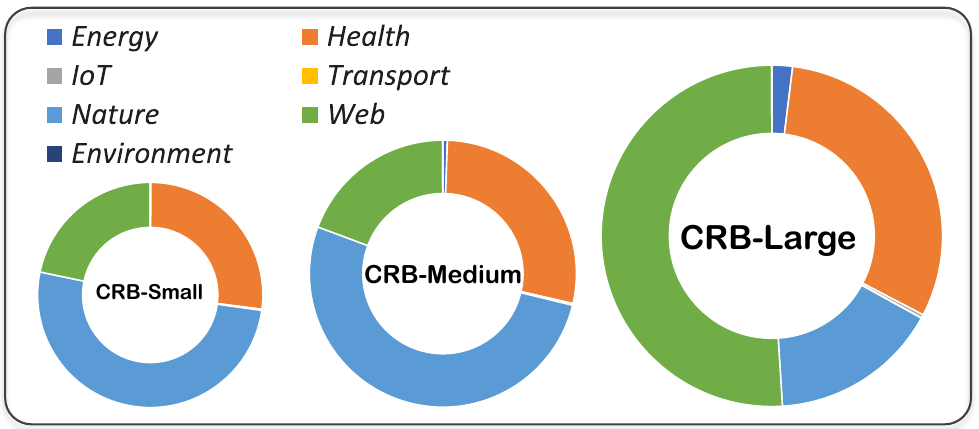}  
    \caption{
     Three Versions of CRB.
    }
    \label{fig:CRBtype}
\end{figure}

Finally, the tree structure supports dynamic updates. When a new sequence \(X_{\text{new}}\) arrives, it is first matched to the nearest prototype using:
\begin{equation}
X_{\text{proto}}^{*} = \arg\min_{X_{\text{proto}}} \| X_{\text{new}} - X_{\text{proto}} \|^2,
\end{equation}
and inserted into the corresponding cluster \(\mathcal{C}_m^{(k)}\). If the cluster exceeds the predefined maximum size \(N\), local re-clustering is triggered within the affected subtree.

This local re-clustering process reassigns the sequences in the overflowed cluster by minimizing the intra-cluster distance:
\begin{equation}
\min_{\{\mathcal{C}_i^{(k)}\}} \sum_{X_j \in \bigcup \mathcal{C}_i^{(k)}} \left\| X_j - C_{i(j)}^{(k)} \right\|^2,
\end{equation}
where the optimization is restricted to the current subtree, and \(C_{i(j)}^{(k)}\) denotes the centroid of the sub-cluster assigned to \(X_j\). This ensures that updates remain computationally tractable and localized.

The above update strategy allows the index to evolve incrementally over time without full reorganization, and is inspired by dynamic clustering methods in data streams~\cite{cao2006density, aghabozorgi2015time}.

\begin{figure}[!h]
    \centering
    \includegraphics[width=0.48\textwidth]{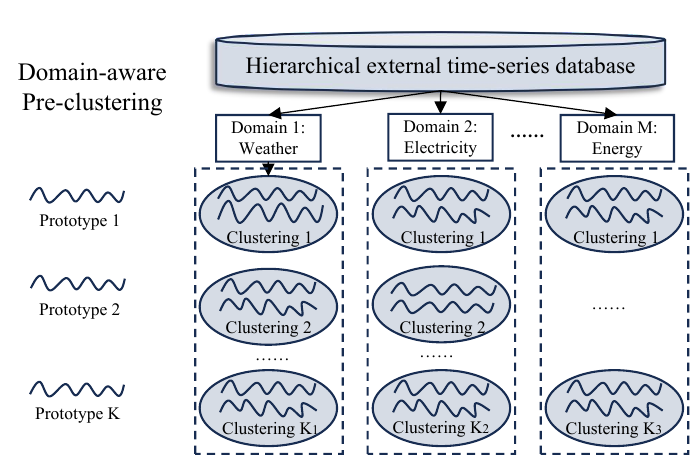}
    \caption{Tree-shaped hierarchical series organization with pre-clustering. Prototypes represent cluster centroids.}
    \label{fig:cluster}
\end{figure}
\label{app:tree}

\section{Language Coherer for LLM based TSPMs}
\label{app:language_coherer}
\begin{figure*}[ht]
    \centering
    \includegraphics[width=0.99\textwidth]{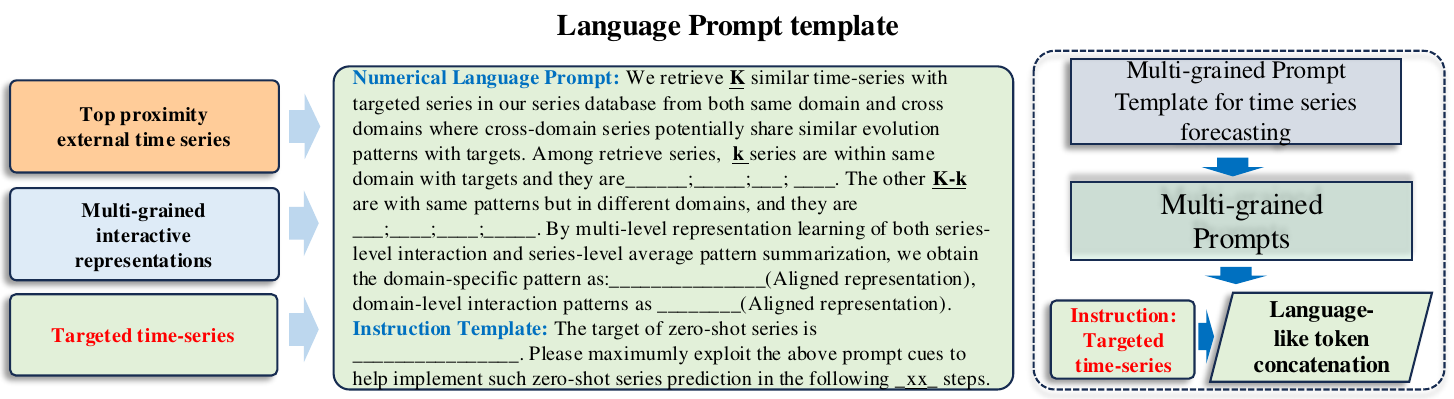}
    \caption{Prompt construction framework for LLM-based forecasting.}
    \label{fig:prompt}
\end{figure*}

Unlike numerical models, language models operate exclusively on text. Direct integration of numeric features is thus infeasible due to modality mismatch. To bridge this gap, we transform the MSIL-derived representations (i.e., \(\mathbf{P}_{\text{int}}\), \(\mathbf{P}_{\text{avg}}\), and \(\mathbf{T}^{\text{norm}}\)) into structured textual summaries (see Figure~\ref{fig:prompt}). These summaries are combined with an instruction prompt to guide the language models in generating forecasting outputs.

The transformation process involves converting numerical features into a format that can be understood by language models. This is achieved by creating textual summaries that capture the essential characteristics of the retrieved time series. The summaries are then combined with an instruction prompt that provides context and guidance for the language model to generate accurate forecasts.

This approach ensures that the language model can effectively leverage the retrieved knowledge, even though it operates on a different modality. By converting numerical features into structured text, we enable seamless integration with the language model, allowing it to generate more accurate and reliable forecasts.

\section{Training Dataset Details}
\label{app:train_datasets}
\begin{figure}[!h]
    \centering
    \includegraphics[width=0.48\textwidth]{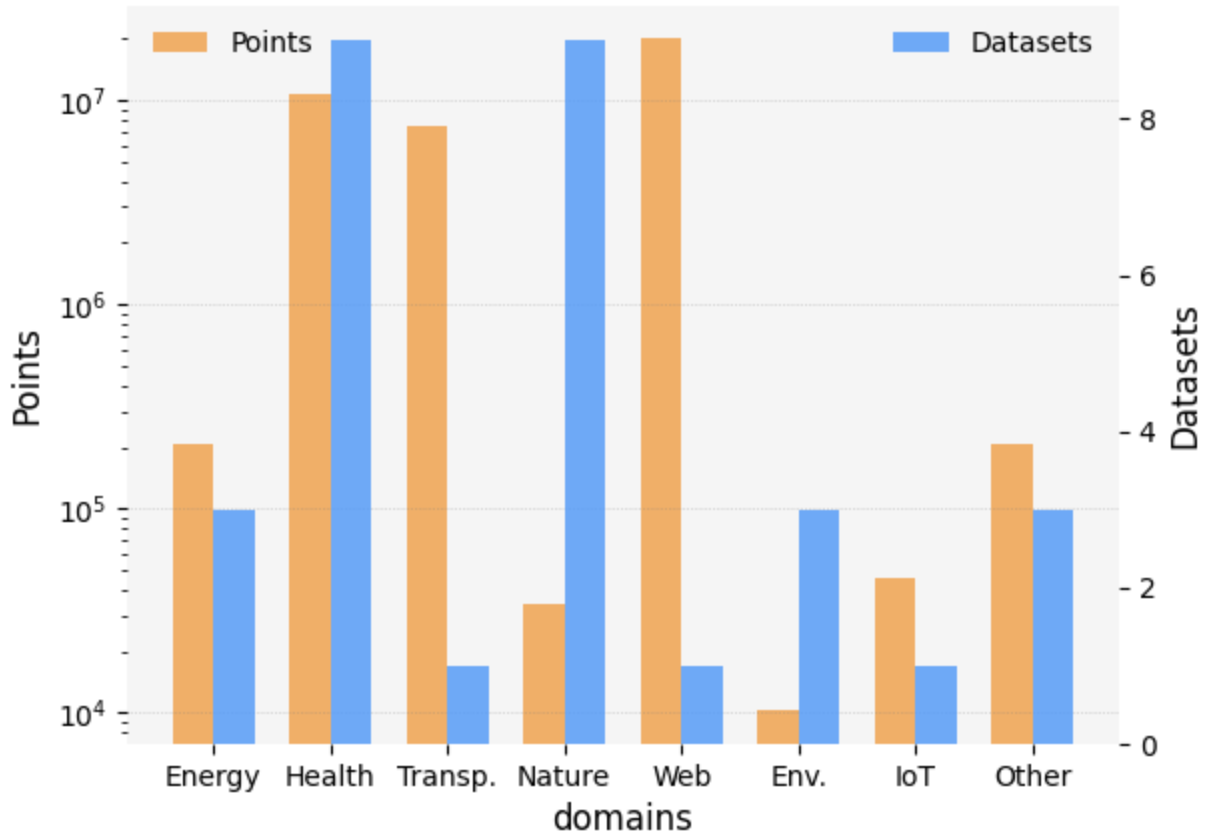}
    \caption{Detailed information of the training set.}
    \label{fig:train_datasets}
\end{figure}

Detailed information of the training set is shown in Figure~\ref{fig:train_datasets}. The x-axis denotes domains. In addition to the seven core domains (Table~\ref{tab:CRB-Large}), extra domains are included to enhance generalization. The y-axis shows the number of time points and datasets per domain.

\section{Additional Experimental Analyses}

\subsection{Complexity Analysis}
\label{app:complexity}
We conduct a detailed empirical analysis and present a comparison of the complexity and efficiency of the QuiZSF$_{T}$ in Figure~\ref{fig:analysis} (a). The comparison is mainly carried out from two aspects: the model size (in mebibytes, MiB) and the CPU inference time per batch (s/iter).
QuiZSF$_{T}$ shows a notable performance in both metrics, with a clear advantage over most comparative models, falling just slightly short of TTM$_B$. Upon further investigation, it is found that this is due to the introduction of learnable modules in the retrieval and feature extraction process based on Retrieval-Augmented Generation (RAG). However, it is worth mentioning that these modules introduced by QuiZSF$_{T}$ are of a lightweight design, with very few additional parameters. Moreover, the retrieval and feature extraction processes rely on dot product calculations, which are highly efficient and do not significantly extend inference time.
The empirical results clearly indicate that while the introduction of QuiZSF$_{T}$ brings certain memory and time overheads, these overheads are within an acceptable range, and at the same time, the model's performance is significantly enhanced. We further believe that by properly adjusting the hyperparameters in the retrieval process and the learnable weights, the number of model parameters can be reduced, thus further optimizing the model.
\subsection{Hyperparameter Study and Analysis}
\label{app:hyper}
\begin{figure}[!h]
    \centering
    \includegraphics[width=0.48\textwidth]{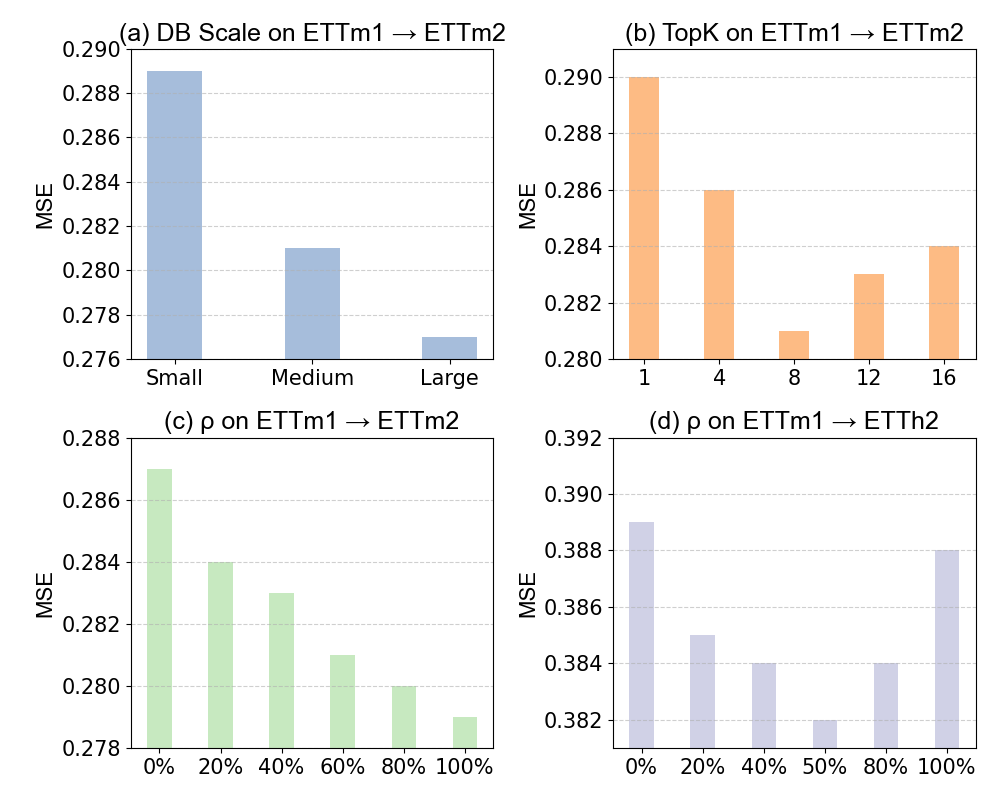}  
    \caption{
     Hyperparameter analysis.
    }
    \label{fig:canshu2}
\end{figure}

We specify three key hyperparameters to explore how to achieve their optimal performance on QuiZSF$_{L}$. Firstly, "Database Scale", which includes three scales of CRB, namely \{Small, Medium, Large\}. Secondly, it is the number of time series retrieved from the auxiliary sequence dataset, with values in the range of $K = \{1, 4, 8, 12, 16\}$. Thirdly, it is the proportion of retrieved time series that are in the same domain as the target series, that is, the Local prototype $\rho$, with values of $\{0\%, 20\%, 40\%, 60\%, 80\%, 100\%\}$. Due to space limitations, the first two experiments are only elaborated in the ETTm1 $\rightarrow$ ETTm2 task, while the last experiment is described in both the ETTm1 $\rightarrow$ ETTm2 and ETTm1 $\rightarrow$ ETTh2 tasks, as shown in Figure~\ref{fig:canshu2}.

The CRB\_Small scale performs worst (Figure \ref{fig:canshu2} (a)). As the scale decreases, the external knowledge it provides declines, leading to poor performance. This partly verifies the scaling law \cite{kaplan2020scaling} in time series.
In the hyperparameter experiment for the retrieved number K, TopK $= 8$ yields the best results. Retrieving more sequences may introduce more noise, while fewer sequences carry less information (Figure \ref{fig:canshu2} (b)).
In the hyperparameter experiment for the Local prototype $\rho$, two tasks show different trends. A too-high proportion of the same domain limits data feature diversity, over-emphasizing single-dimension features. The model performs well on samples fitting this feature (Figure \ref{fig:canshu2} (c)), but poorly on non-matching samples due to the lack of auxiliary correction from other dimensions, resulting in a polarized outcome (Figure \ref{fig:canshu2} (d)).
To balance performance and efficiency, we select CRB\_Medium, set Top$K$ to 8, and set the Local prototype $\rho$ at $60\%$.

\subsection{Case Study}
\label{app:case}
In order to demonstrate how the retrieval sequence improves the prediction effect, we conduct an intuitive analysis of the intermediate results. Taking the target sequence of the ETTh2 dataset as an example (as shown in Figure~\ref{fig:sample}), we screen out the Top-8 sequences that are closest to the target sequence through hybrid and hierarchical time-series retrieval and marked their similarities. These sequences have similar patterns and evolution models as the target sequence. Subsequently, we update the sequence representation of the target sequence by combining the auxiliary sequence with the target sequence through MSIL and input it into the LLM after generating prompts. We visualized and compared the output of the LLM with RAG and the output w/o RAG. 

The results show that RAG can reveal the average pattern of the retrieval sequence, making the prediction results smoother and avoiding overfitting; while the output w/o RAG fluctuates more and contains more inaccurate details. This indicates that RAG effectively suppresses the time-series hallucination of the LLM. Our analysis enhances the interpretability of the model, deepens the understanding of zero-shot forecasting, and highlights the contribution of RAG in enhancing prediction.
\subsection{Experiments compute resources}
\label{app:resource}
All experiments were conducted on a single NVIDIA A100 GPU with 40GB memory. 

\section{Limitations and Future Work}
\label{app:limition}

QuiZSF, while advancing zero-shot time series forecasting, faces limitations: its effectiveness relies on pre-trained model quality, which is limited in data-scarce domains; large-scale retrieval efficiency remains a challenge, potentially addressable with sparse indexing; and multi-granularity sequence learning for cross-pattern transfer is needed to enhance generalization across heterogeneous streams, ultimately strengthening QuiZSF and contributing to scalable retrieval-augmented AI systems for diverse Web environments.

\end{document}